# Cost Scales with Change, Not Corpus Size: Incrementally Maintaining an Evolving Semantic Substrate

Yusuke Takahashi
*Asia AI Institute,*
*Faculty of Data Science*
*Musashino University*
*Tokyo, Japan*
*ORCID: 0009-0006-8351-2280*

Kyle Wild
*Endgame Labs, Inc.*
*San Francisco, USA*
*Asia AI Institute*
*Musashino University*
*Tokyo, Japan*
*ORCID: 0009-0001-6918-3197*

Asako Uraki
*Asia AI Institute,*
*Faculty of Data Science*
*Musashino University*
*Tokyo, Japan*
*ORCID: 0009-0006-8412-1804*

***Abstract*—Retrieval-augmented and agentic question-answering systems increasingly re-derive the meaning of a corpus at query time. Put plainly, instead of re-deriving what a corpus means on every question, the work is done once when a document arrives and is thereafter merely consulted—a compiler, not an interpreter, of meaning. An alternative is to compile that meaning once, at ingest time, into a compact, queryable semantic substrate and maintain it as the corpus evolves. The central objection is maintenance cost: rebuilding a truncated singular value decomposition (SVD) on every change appears prohibitive, and a change of embedding model seems to force a full re-embedding. We argue and show empirically that maintenance cost scales with the amount of change, not corpus size. On a controlled synthetic pilot (dimension 256, rank 32, a corpus grown from 3,000 to 9,000 documents over 50 update events), incremental low-rank updates were 33.7 times cheaper per update than full re-SVD and 23.8 times cheaper cumulatively, while the incremental subspace tracked the full recomputation to within floating-point precision (maximum principal-angle drift below $10^{-11}$ degrees; recall@10 = 1.0). An orthogonal Procrustes virtual axis update recovered 0.95 mean cosine to truly re-embedded vectors by re-embedding only about 10% of the corpus. The results support maintaining, rather than repeatedly reconstructing, a semantic substrate.**

***Index Terms—semantic substrate; incremental SVD; retrieval-augmented generation; embedding drift; orthogonal Procrustes; latent semantic indexing; knowledge bases***

## I. Introduction

Modern question-answering over documents commonly follows a query-time pattern: when a question arrives, the system retrieves passages and a language model reconstructs the relevant meaning from raw text on the spot. Retrieval-augmented generation popularized this retrieve-then-generate paradigm [1], and recent systems even let a model translate natural-language questions into executable plans over heterogeneous data [2]. We call this query-time semantic reconstruction (QSR). Increasingly, the same role is played by retrieval-oriented Model Context Protocol (MCP) servers [3]—the emerging open standard by which assistants reach external data sources at run time.

QSR is flexible, but it repeats semantic work on every query and offers limited control over consistency, provenance, and cost as query volume grows; it is also sensitive to how much context a model can effectively use [4]. An alternative is to perform the semantic work once, at ingest time, compiling the corpus into a compact, queryable semantic substrate that later queries consult—ingest-time semantic compilation. By analogy to language implementation, QSR is an interpreter that re-translates on every run, while ISC is a compiler that translates once and reuses the result (Fig. 1).

Applying this idea to learned semantic representations, however, raises a practical objection: as the corpus changes, keeping the substrate current seems expensive, and upgrading the underlying embedding model appears to require re-embedding everything. We show the objection is largely misplaced.

***Contributions.*** (i) We frame substrate maintenance as the key cost question and argue that, with incremental low-rank updates, cost scales with change rather than corpus size. (ii) On a controlled synthetic pilot we show incremental updates are an order of magnitude cheaper than full reconstruction while tracking it to floating-point precision. (iii) We show an orthogonal Procrustes virtual axis update can absorb an embedding-model generation change by re-embedding only a small anchor fraction. The novelty is thus not a new factorization algorithm but the maintenance discipline itself: framing substrate upkeep as the central cost question, unifying incremental SVD updates and Procrustes migration into a single change-driven procedure, and measuring that the resulting cost scales with change rather than corpus size.

The remainder of this paper is organized as follows: Section II reviews related work; Section III defines the semantic substrate and its maintenance and migration procedures; Sections IV–VI present results, discussion, and conclusions.

## II. Literature Review

### A. Semantic spaces and orthogonality

Vector space models represent documents and queries as high-dimensional vectors ranked by cosine similarity [5]. Latent semantic indexing applies a truncated SVD so related terms map to nearby directions in a low-rank space whose singular vectors are orthogonal [6]. A mathematical model of meaning built an explicitly orthogonalized semantic space on a basic-vocabulary basis, treating orthogonality as the property that lets cosine behave as a semantic distance [7]. We treat them as related precedent, not a prerequisite.

Accepted for publication in the 2026 International Electronics Symposium (IES), Yogyakarta, Indonesia, August 1–3, 2026. This is the authors' accepted version; the final published version will appear in IEEE Xplore.

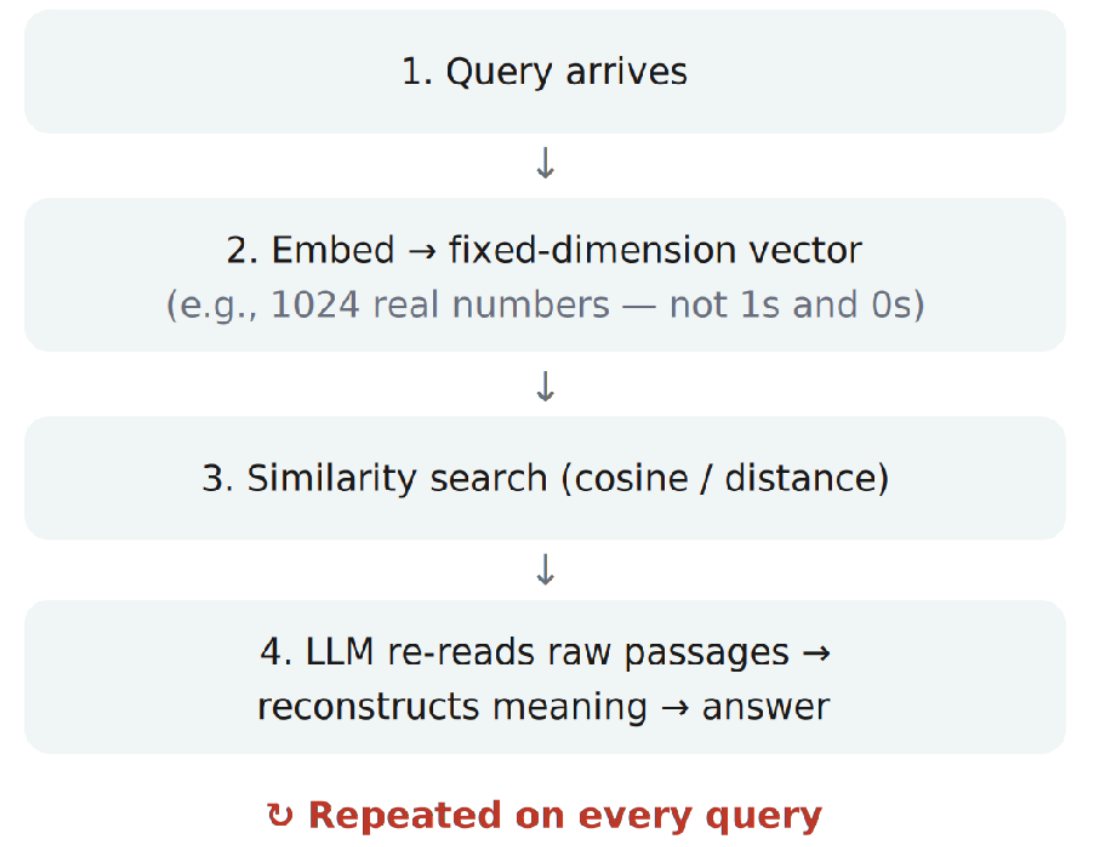

Fig. 1. QSR reconstructs meaning on every query; ISC compiles meaning once at ingest and maintains it incrementally, so a query becomes a cheap lookup.

### *B. Embedding geometry and anisotropy*

Contextualized embeddings are anisotropic: vectors occupy a narrow cone, inflating cosine similarities between unrelated items [8]. Post-hoc corrections—flow-based normalization [9], contrastive training [10], and cluster-based isotropy enhancement [11]—can be read as partial, retrofitted re-orthogonalization, and benchmark studies show embedding quality varies sharply across tasks and models [12]. This motivates compiling and, where useful, re-orthogonalizing a substrate rather than relying on raw embeddings; their limitation is that the corrections are global and static, not maintained as a corpus evolves.

### *C. Dense retrieval and retrieval-augmented generation*

Dense retrieval learns query and passage encoders for nearest-neighbor search [13], with late-interaction [14] and fusion-in-decoder [15] variants, and self-reflective retrieval that decides when to retrieve [16]. These define the QSR baseline: meaning is reconstructed per query; production file-search features and retrieval-oriented MCP servers [3] instantiate it[1]. Their limitation, for our purposes, is that none persists a compiled, maintainable semantic structure; each query re-pays the semantic cost, and an embedding-model upgrade invalidates the index.

### *D. Indexing, maintenance, and migration*

On the read side, approximate nearest-neighbor indexes make substrate lookups practical [17], [18], [19]. On the write side, incremental and online SVD update a factorization as rows are added without recomputing from scratch [20], [21], and nested representations show a learned space can be safely truncated for rank control [22]. Migration between spaces is standard: the orthogonal Procrustes problem has a closed-form solution [23], and backward-compatible representation learning targets model upgrades directly [24]. A complementary symbolic line stores compiled meaning as reusable knowledge objects [25]; the substrate maintained here is its continuous counterpart. The gap this paper fills is that these write-side tools have not been combined and measured for the specific goal of keeping a compiled semantic substrate current—at a cost set by change—across both content edits and model upgrades.

## III. Methods

### *A. Preliminaries*

We summarize the minimal intuition needed for this section. An embedding represents a document's meaning as a list of a few hundred numbers (a vector); documents with similar meanings point in similar directions. A corpus is then one large table (a matrix) whose rows are these vectors. The singular value decomposition (SVD) is the standard tool that reorganizes this table into a small number of principal semantic axes; keeping only the top k axes yields a much smaller representation (a low-rank approximation) that preserves the gist. The semantic substrate in this paper is exactly this small set of axes. When the axes are mutually orthogonal (non-overlapping), the angle between vectors (cosine similarity) behaves as an honest measure of semantic closeness. Finally, Procrustes alignment finds the single rotation that best overlays the same point set expressed in two coordinate systems without distorting shapes—like turning a whole room while keeping the furniture's relative arrangement. The subsections below restate this intuition precisely.

### *B. Semantic substrate*

We embed each document into a D-dimensional vector and stack the corpus as $X \in \mathbb{R}^{N\times D}$. The substrate is a rank-k factorization from a truncated SVD,

$$X \approx U_k \Sigma_k V_k^T, \quad V_k^T V_k = I_k ,$$

whose top-k right singular vectors $V_k$ form an orthonormal basis of a low-rank semantic subspace. A query q is projected as $\tilde{q} = q V_k$ and scored by cosine similarity, at $O(Dk)$ per item.

### *C. Incremental maintenance vs. full re-SVD*

When a batch of m documents is added or revised, the naive option recomputes the truncated SVD over the whole corpus (full re-SVD) at roughly $O(NDk)$ per iteration—growing with N. Incremental low-rank updates [20] instead modify $(U_k, \Sigma_k, V_k)$ directly using only the change, at roughly $O(Dk\cdot m + Dk^2)$ per update—independent of N to first order. We periodically re-orthogonalize to bound departure from $V_k^T V_k = I_k$. Intuitively, full re-SVD re-reads the entire library every time a new shipment of books arrives, whereas the incremental update merely shelves the new arrivals and adjusts the catalogue.

### D. Virtual axis update for model changes

On an embedding-model upgrade, old and new vectors are not comparable. We embed a small anchor set under both models, giving M and $\tilde{M}$, and solve the orthogonal Procrustes problem [23],

$$R^* = \mathrm{argmin}_R \ \| MR - \tilde{M} \|_F \ \ \text{s.t.}\ R^T R = I, \quad R^* = WZ^T, \quad M^T\tilde{M} = W\Sigma Z^T,$$

so that applying R* virtually updates the axes and only the small anchor fraction is re-embedded—trading a full rebuild for an alignment problem, complementary to backward-compatible training [24].

### E. Rank control and break-even

The rank k trades fidelity for size; nested-representation training [22] suggests k can be adapted to a target precision. Maintaining a substrate is worthwhile only if its amortized cost beats per-query reconstruction. Precomputation is a spectrum, not a binary: a vector index already amortizes retrieval (embedding and nearest-neighbor search), but not the per-query semantic reconstruction a model performs over raw passages—so $c_q$ reflects that reconstruction, which is precisely what ISC persists and maintains. With ingest cost $c_c$ per document, maintenance cost $c_m$ per change, QSR query cost $c_q$ and lookup cost $c_r$, the break-even read count is

$$R^* = (N \cdot c_c + W \cdot c_m) / (c_q - c_r) .$$

Because incremental maintenance keeps $c_m$ small and independent of N, R* stays low in high-query regimes.

### F. Experimental design

We use a controlled synthetic pilot that isolates maintenance cost and tracking quality from confounds; it is a harness, and a real-corpus study is future work. Each document is a vector $x = B_t c + \varepsilon \in \mathbb{R}^D$ with D = 256. Here $B_t \in \mathbb{R}^{D\times R0}$ (R0 = 40) is a latent semantic basis, orthonormalized by QR, that drifts slowly across events via a small random rotation (perturbation scale 0.03), modeling a corpus whose topics shift gradually; the coefficients $c_i \sim N(0,1)$ are scaled by $1/\sqrt{(1+i)}$, giving a decaying per-factor spectrum so a few factors dominate; and $\varepsilon \sim N(0, 0.15^2 I)$ is off-subspace noise. The substrate rank k = 32 is deliberately below the latent rank R0 = 40, so the substrate is a genuine low-rank approximation, not an exact fit. The corpus starts at N0 = 3,000 documents and grows over 50 events, each appending a batch of m = 120 freshly generated documents, reaching N = 9,000. For the model-migration test we build a separate corpus of 4,000 documents and simulate an embedding-model change as a random orthogonal rotation of that space plus Gaussian noise (scale 0.05). All randomness is seeded (global seed 7; each event's drift seeded deterministically) for exact reproducibility.

### G. Metrics and implementation

We measure per-update and cumulative wall-clock cost; the maximum principal angle between the incremental and full top-k right-singular subspaces; recall@10 of nearest-neighbor retrieval over a fixed set of 200 random queries, relative to full recomputation; and reconstruction error. Full re-SVD recomputes the top-k right singular vectors of the whole corpus (cost $\sim O(N D^2)$); the incremental maintainer accumulates only the new batch into a running D×D second-moment matrix and re-extracts the top-k eigenvectors (cost $\sim O(m D^2 + D^3)$, independent of N). For migration we learn the orthogonal Procrustes map from $k \in \{50, 100, 200, 400, 800\}$ anchor pairs and apply it to the whole corpus without re-embedding. The pilot runs on a single CPU core with standard dense linear algebra; embedding time, shared by both methods, is excluded from timings.

TABLE I. Synthetic-pilot results (D=256, k=32). Incremental maintenance is 23.8–33.7× cheaper than full re-SVD yet matches it (drift below $10^{-11}$ deg; recall@10 = 1.0): cheaper and not lossy.

| Metric | Full re-SVD | Incremental |
|---|---|---|
| Per-update cost (N=9,000) | 283 ms | 8.4 ms (33.7×) |
| Cumulative cost (50 events) | 10.0 s | 0.42 s (23.8×) |
| Max principal-angle drift | — (ref.) | $<10^{-11}$ deg |
| recall@10 vs. full | 1.0 | 1.0 |

## IV. Results

***Cost asymmetry.*** At N = 9,000, incremental maintenance cost 8.4 ms per update versus 283 ms for full re-SVD—33.7× cheaper—and 23.8× cheaper cumulatively over the 50 events (Table I). Incremental per-update cost stayed flat as the corpus grew, whereas full re-SVD rose with N (Fig. 2).

***Subspace tracking.*** Across all events the maximum principal-angle drift between the incremental and full subspaces stayed below $10^{-11}$ degrees and recall@10 was 1.0 (Fig. 4); the incremental reconstruction error matched the full baseline to working precision.

***Virtual axis update.*** An orthogonal Procrustes map from a small anchor set recovered 0.95 mean cosine to truly re-embedded vectors while re-embedding only about 10% of the corpus (Fig. 5).

***Scaling.*** The full re-SVD curve rises with N, whereas the incremental curve is flat, set by the batch size m rather than N; the cumulative gap therefore widens as the corpus grows (Fig. 3).

## V. Discussion

The results reframe the usual objection. Maintenance is dominated by change, not corpus size: a large but stable corpus is cheap to keep current; only churn incurs cost. A maintained substrate is therefore attractive precisely in the high-volume, slowly-changing regimes where QSR [1], [13] repeats the most work, and the break-even count R* stays low.

The virtual axis update softens the most feared maintenance event—an embedding-model upgrade—by converting a full re-embedding into a small alignment problem, complementing backward-compatible training [24]. The substrate also complements language-model-driven query systems [2] and is the geometric counterpart to symbolic knowledge-object memory [25], offering cheap nearest-neighbor recall alongside explicit, inspectable facts.

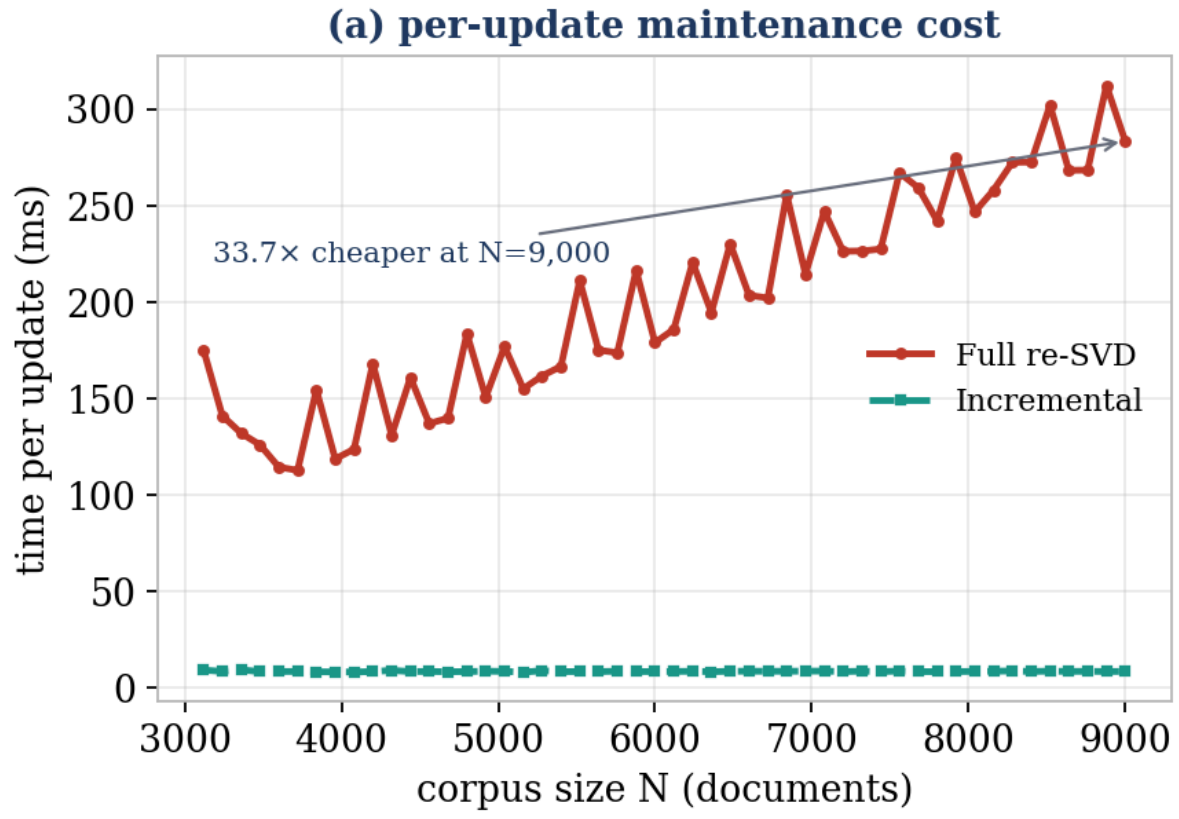

Fig. 2. Per-update maintenance cost vs. corpus size. Incremental updates stay flat (set by the batch size m) while full re-SVD rises with N: maintenance cost scales with the amount of change, not corpus size (Contribution i).

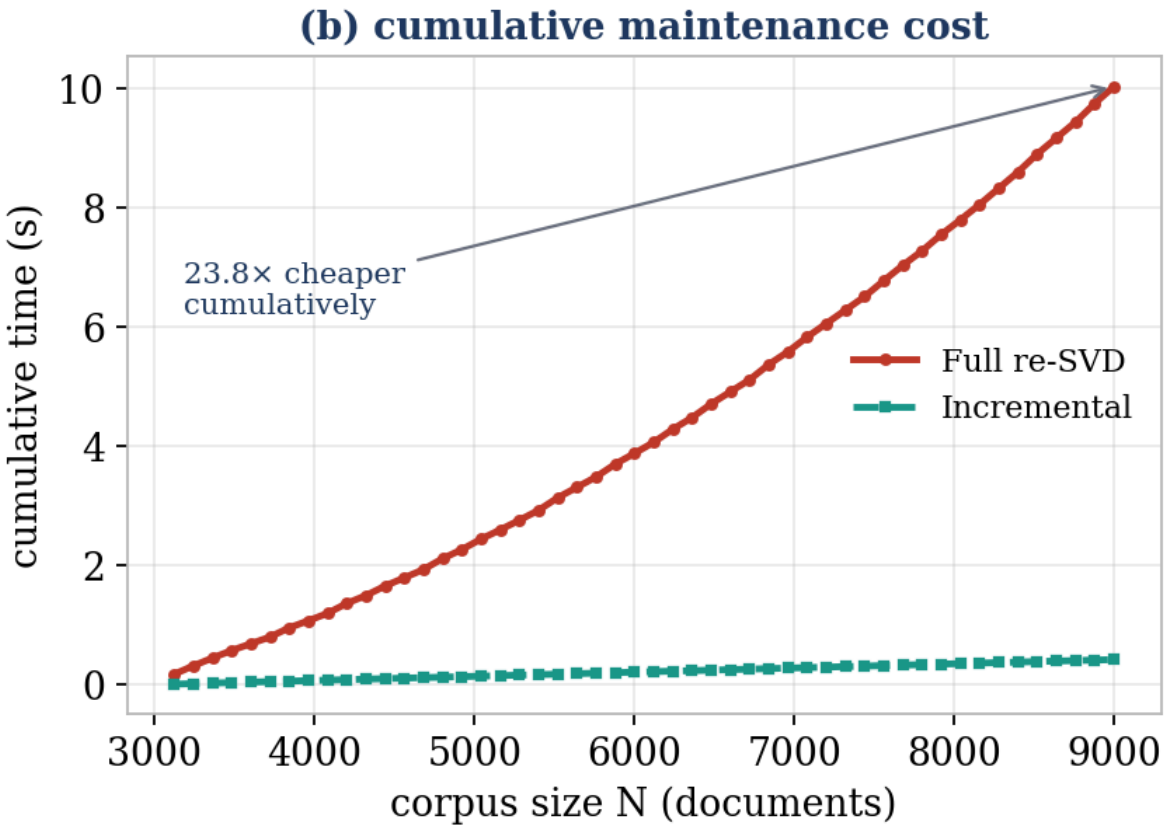

Fig. 3. Cumulative maintenance cost over the 50 update events; the gap widens as the corpus grows (23.8× cumulatively).

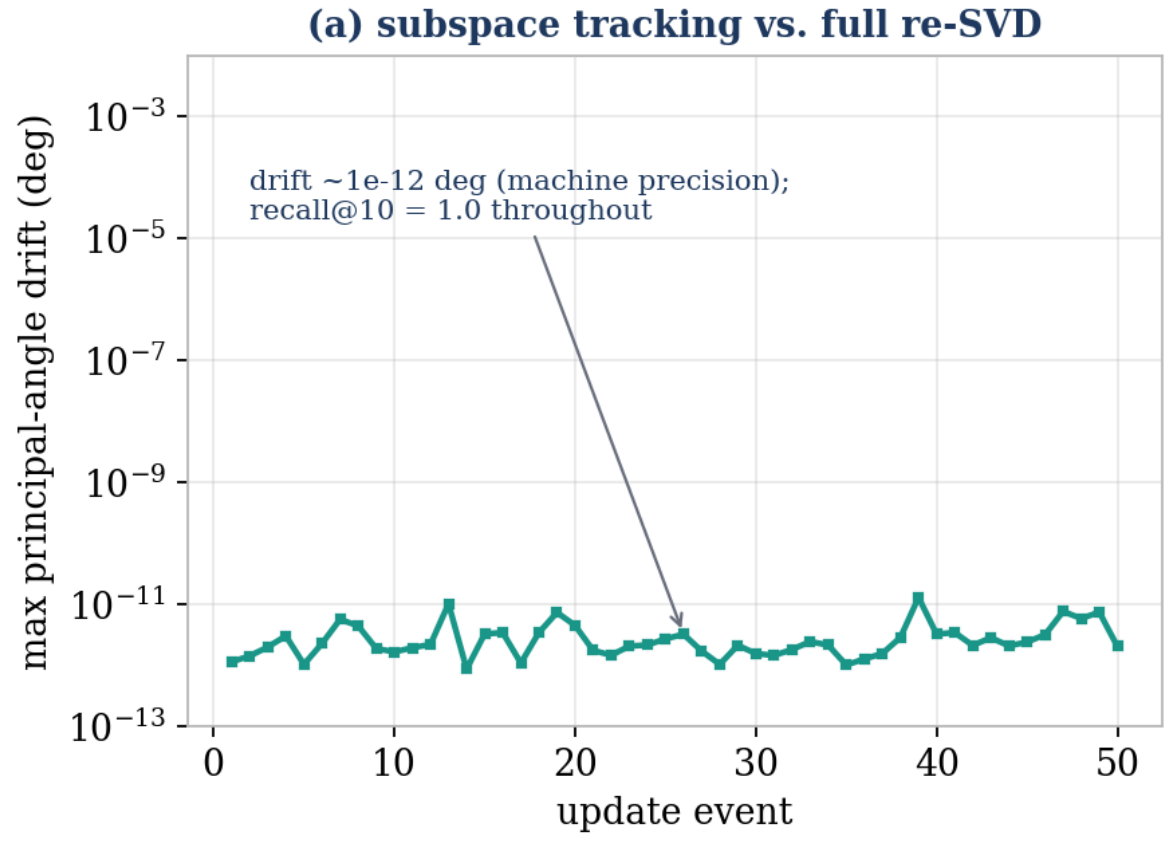

Fig. 4. Maximum principal-angle drift between the incremental and full subspaces over update events (log scale). Below $10^{-11}$ degrees throughout, with recall@10 = 1.0: the cheap incremental substrate is not a lossy shortcut—it matches full re-SVD to floating-point precision (Contribution ii).

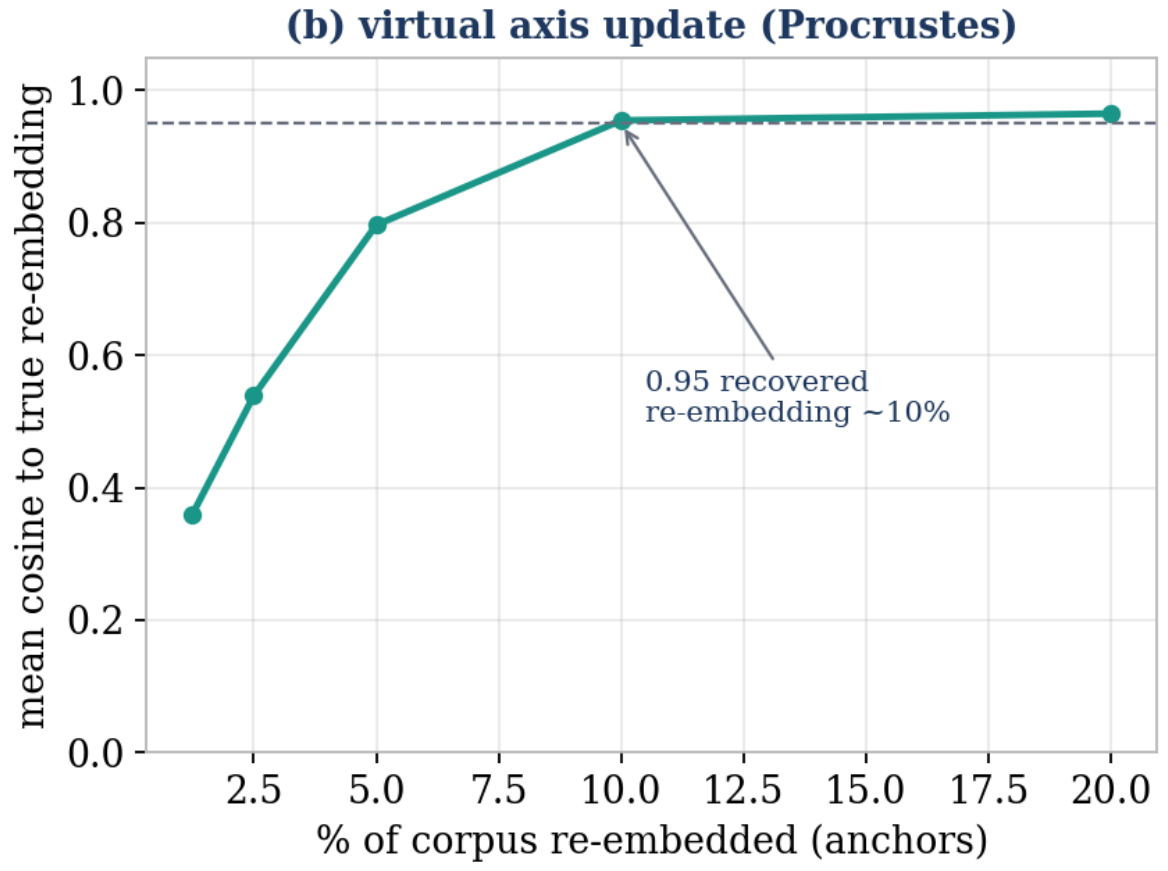

Fig. 5. Procrustes virtual axis update: re-embedding only ~10% of the corpus recovers 0.95 mean cosine to truly re-embedded vectors, so an embedding-model upgrade need not force a full re-embed (Contribution iii).

For end-user assistants the practical implications are concrete. Repeated questions over the same corpus become faster and cheaper, because the ingest-time work is reused rather than re-paid on every query. Answers can be made more auditable, and persisting compiled facts can reduce drift over long interactions. We therefore see ISC not as a replacement for QSR but as a complementary mode that assistants such as those above could adopt for heavy, repetitive corpus processing. Because MCP is a delivery interface rather than a retrieval strategy, a maintained substrate can itself be served behind an MCP server, letting agents reach a compiled, always-current semantic layer through the same open standard they already use for query-time retrieval. Evaluating answer quality is, moreover, more than a matter of truth alone; a multi-dimensional framework for AI-output evaluation [26] complements this auditability argument.

***When is compiling not worth it?*** If a corpus is small, highly volatile, or queried rarely, the read count may never reach R*, and per-query reconstruction remains preferable. The claim is not that compiling always wins, but that its dominant maintenance cost is change.

***Limitations.*** The pilot is synthetic and uses an idealized incremental update, so the absolute speedups and the exact-tracking result are best-case bounds; rank-one streaming variants and noisy real corpora will introduce bounded drift, and Procrustes recovery depends on anchor representativeness. Real embedding APIs and a live revision stream are needed for external validity.

## VI. Conclusion

An evolving semantic substrate can be maintained, rather than repeatedly reconstructed, because maintenance cost scales with change, not corpus size; and an embedding-model change can be absorbed by a virtual axis update instead of a full rebuild. A synthetic pilot supports both claims. Future work will measure on real corpora with production embedding APIs and a live revision stream, and study the complementary symbolic substrate—e.g., knowledge objects [25]—for provenance and adherence.

Concretely, we will replay a timestamped Wikipedia revision stream against production embedding APIs and compare against existing vector-database re-indexing practice, extending the present pilot to real-world retrieval and RAG deployments.

## Acknowledgment

Part of this work used computational infrastructure provided by Endgame Labs, Inc.

[1] Run-time retrieval features cited as QSR examples: OpenAI file search; Google Gemini File Search; Anthropic Claude file uploads and web search.